# Bootstrapping Deep Neural Networks from Approximate Image Processing Pipelines

Kilho Son, Jesse Hostetler, Sek Chai
SRI International
sek.chai@sri.com

## ABSTRACT

Complex image processing and computer vision systems often consist of a processing pipeline of functional modules. We intend to replace parts or all of a target pipeline with deep neural networks to achieve benefits such as increased accuracy or reduced computational requirement. To acquire a large amount of labeled data necessary to train the deep neural network, we propose a workflow that leverages the target pipeline to create a significantly larger labeled training set automatically, without prior domain knowledge of the target pipeline. We show experimentally that despite the noise introduced by automated labeling and only using a very small initially labeled data set, the trained deep neural networks can achieve similar or even better performance than the components they replace, while in some cases also reducing computational requirements.

## 1 Introduction

Notwithstanding the recent popularity of deep neural networks for image processing and computer vision (IP/CV) applications, there are a great many IP/CV systems already "in the wild" that are based on more conventional algorithms. These often take the form of a pipeline of components that perform a sequence of transformations to yield the final result. A functioning processing pipeline represents a significant engineering investment, and it may have been validated through years of service so that there is a high degree of confidence in its performance. Such an investment is not to be discarded lightly. On the other hand, the rise of "big" data and the demand for greater capabilities may make it necessary to improve the efficiency, scalability, or accuracy of the system.

An IP/CV pipeline is typically a combination of both engineered and learned components. Developing the pipeline in the first place requires substantial engineering effort in choosing the right methods, tuning their parameters, and integrating them into an end-to-end system. Revisiting the pipeline to make changes or improvements requires thorough understanding by domain experts or developers. This may be costly, especially if the system is old and engineering knowledge has been lost.

Thus, it is desirable to improve the existing processing pipelines without understanding the complex system. That is, we need to optimize the target pipeline as a combination of black boxes where we can only measure inputs and outputs of each part.

For an increasing number of IP/CV tasks, deep neural networks (NNs) represent the current state-of-the-art. Thus, an obvious approach to improving an IP/CV system is to replace parts of the processing pipeline with state-of-the-art neural networks. This may bring an immediate benefit in terms of increased accuracy. It also enables the NN portion of the pipeline to exploit advancements algorithms or specialized hardware developed within the greater field of deep learning.

One of the major obstacles in adopting deep neural network methods is that a large amount of labeled training data is required to avoid overfitting. Much of the training data today have been curated and labeled manually at great cost in human resource. The learned components of IP/CV pipelines will usually be simpler than a deep NN, and thus the data used to train the pipeline may not be adequate for training a NN to replace it. The training data may have been lost or become difficult to obtain since the pipeline was developed. Some processing pipelines may have no learned components, and thus no training data.

In this paper, we propose to optimize existing IP/CV pipelines using deep NNs to acquire better performance with reduced or similar computational requirement. This optimization is done by replacing either the whole target pipeline or some important parts of it with deep neural networks. We exploit good characteristics of deep neural networks: 1) effectiveness in modeling patterns between inputs and outputs of a target system without expert knowledge of the target system, 2) robustness that possibly rejects noise and outliers in the training data and 3) efficiency thanks to optimal software and hardware for deep NNs. We acquire the large amount of training data needed for deep neural networks by bootstrapping the NN model from the target pipeline in a semi-supervised fashion. From our experiments with image processing and computer vision pipelines, we observed that despite the noise introduced by automated labeling, the trained neural networks achieve similar or even better performance than the components they replace, while in some cases also reducing computational requirements. Although we limit our applications to IP/CV pipelines, our proposed frameworks can be applied general signal processing applications.

To train the neural networks, we propose to use noisy labels generated from the existing target pipelines as well as data with ground truth labels if available. In this way, we can remodel the target pipeline without full knowledge of the target parts of the pipeline. After this optimization, the system is expected to improve either efficiency or performance, or both efficiency and performance depending on objective of the optimization. This could be applied not only particular parts of the pipeline but also all the pipeline.

## 2  Related Work

Our work can be considered an approach to approximate computing [3,13], in which a target function is approximated by a surrogate that has less compute but introduces inaccuracy. In IP/CV, some level of inaccuracy is often tolerable, due to the limits of human perception and the lack of a clearly delineated "correct" answer [19]. Approximation can be introduced at the hardware level, such as by using approximate adder circuits, e.g. [9], or at the software level by restructuring the algorithm.

Our work can also be seen as an application of the semi-supervised learning paradigm [22], where the learner is given both labeled and unlabeled training data. We take a bootstrapping or "self-supervised" approach [21, 16], using elements of the processing pipeline as surrogate models to label the unlabeled examples. The imputed labels will contain errors, and thus techniques for learning from noisy labels [14, 15] are also relevant. Several works have shown that neural networks can be trained successfully based on noisy labels (e.g. [16, 18, 2, 6]).

## 3  Proposed Workflow

Figure 1 illustrates our proposed workflow for deep NN approximation of existing target software pipelines. We describe this workflow with an example program consisting of functional modules that is identified for approximation (e.g. the target pipeline in green). For example, the functional module may consist of an SVM-based image classification that we approximate using deep NN. The inputs and outputs to the target pipeline are generated as the dataset to train a neural network as an approximation. The goal is acquiring better algorithmic performance with reduced or similar computational requirement.

We describe the workflow in more algorithmic detail. Suppose we wish to replace a component P of a processing pipeline. The goal is to optimize the pipeline for better performance and/or reduced computational cost by replacing P with a neural network F that performs the same function. We treat P as a black box mapping inputs to outputs, $P : X \rightarrow Y$. If P is a model learned from data, it may be possible to learn F from the same training data. However, in many cases this is not feasible. It may be that there was sufficient training data to learn P, but not enough to learn F. If some time has passed since the development of the processing pipeline, the training data for P may have been lost.

We assume that we can acquire a sufficiently large set of *unlabeled* inputs $U \subset X$. In some cases, we may also have access to a set $L \subset (X \times Y)$ of *labeled* data, but we assume that $|L| \ll |U|$. Our proposed approach is to compute a set $S = \{(x, P(x)) : x \in U \}$ of *imputed* labeled examples using P as a surrogate model to perform the labeling. This is an application of the idea of self-supervision from semi-supervised learning.

By generating large amounts data labeled by the target pipeline, we can have access to enough training data to prevent a complex deep neural network from overfitting to a small number of datum with ground truth labels. On the other hand, the labels in the set S are likely to be noisy due to imperfect performance of the model P used to generate them. When the pipeline component P is computationally expensive, a small loss of accuracy may be acceptable to reduce computational power requirements. In the next section, we demonstrate experimentally that not only can replacing P with a neural network F significantly reduce computation in some problems, the accuracy F may be greater than that of P even though F was trained on noisy labels generated by P.

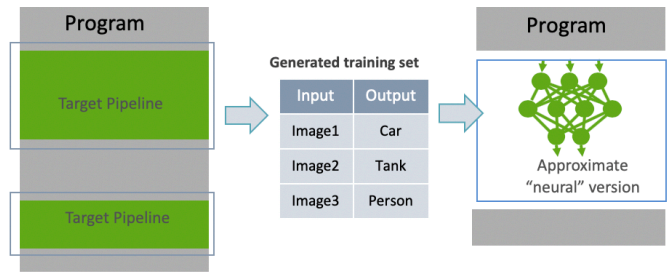

Figure 1: Our proposed workflow enables generation of a deep Neural Network (NN) as an approximation of a target pipeline. Input/output training data set is automatically curated from conventional (target) pipeline to train the deep NN.

## 4  Experiments

We have evaluated the proposed framework with image processing and computer vision pipelines, specifically image denoising and image classification. Image denoising and image classification are targeting goal as they are, or role parts of system pipelines. Although our proposed framework is able to optimize any start and end point of the target pipeline once we can measure inputs and corresponding outputs, we selected image denoising and image classification pipeline for the experiments because we can simply evaluate performance and efficiency of algorithms with available test data with ground truth labels. Although we limit our applications to image processing and computer vision in this paper, we note that our proposed method can optimize general signal processing pipeline.

In this experiment, we assumed that a target pipeline that we want to optimize is a black box. That is, the target pipeline is given where we can only probe inputs and corresponding outputs. Data with ground truth labels (inputs and corresponding ground truth outputs) of the pipeline may be partially available or not. As proposed in the Section 3, we feed unlabeled data to the target pipeline and generate labels which is used for training deep neural networks. If data with ground truth is available, we also use this data along with the noisy data. Our intent is to show the performance of the deep NN trained with different ratio of ground truth labels and generated labels (from the target pipeline). As such, the amount of training data is the same, and we sweep over the different ratio of ground truth versus generated labels.

### 4.1 Image Denoising

For image denoising experiment, we have used BSD300 [12] which consists of 300 images. Similar to the experiment setup by Mao *et al.* [11], 50x50 pixel-patches are randomly selected in random 200 images generating 20000 number of training data. 4000 number of testing data are generated from random patches in the other 100 images. To generate input noisy images, a value from Gaussian distribution with mean 0 and variance 20 is added to each pixel independently followed by truncation when the pixel value is below 0 or above 255.

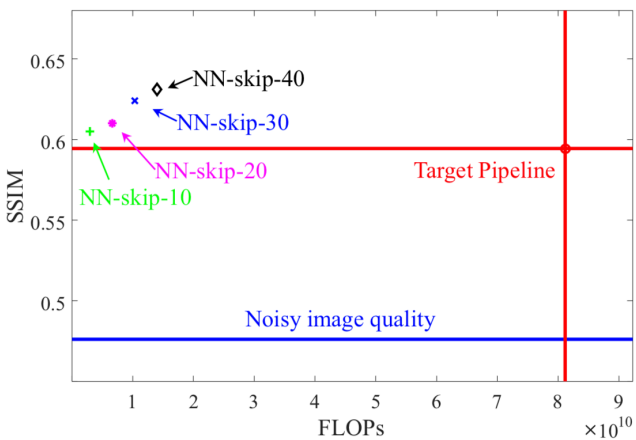

Figure 2: Remodeling a target image denoising pipeline with deep neural networks: The deep neural networks are trained solely by data labeled by the target pipeline. We note that all the deep neural networks improve performance with reduced computational requirements. FLOPs is a number of multiplication and addition which is corresponding to computational requirement.

To measure the quality of the images given ground truth images, we used a metric, Structural Similarity (SSIM) index [20]. SSIM is a perception-based model that considers image degradation as perceived change in structural information, while also incorporating important perceptual phenomena, including both luminance masking and contrast masking terms [20]. The SSIM metric ranges from 1 to 0, where higher value means better quality of images.

The target image denoising pipeline formulates denoising problem as Conditional Random Field (CRF) optimization and optimize the CRF objective via the α-expansion and α − β-swap algorithms [1, 5]. The target pipeline is unsupervised where the required parameters for the pipelines are all predefined by the algorithm developer. To optimize this pipeline, we have used feed-forward autoencoder neural network with symmetric skip connections [11] with 10, 20, 30, 40 layers, refer to NN-skip-10, NN-skip-20, NN-skip-30 and NN-skip-40 respectively.

We simulated that noisy input images are available without ground truth labels for training. As proposed, we trained the deep neural networks (NN-skip-10, NN-skip-20, NN-skip-30 and NN-skip-40) with all the training data labeled by the target pipeline where the labels are possibly incorrect. Figure 2 shows image quality (SSIM) of input noisy test images and their qualities after denoising using each method along with the computational requirements. FLOPs are modeled as a number of additions and multiplications of an algorithm. When we use the target pipeline, the quality of test images is improved from 0.476 to 0.594. Interestingly, when we use the deep neural networks learned by the training data labeled by the target pipeline, the performance on test images are more improved (0.605, 0.610, 0.624 and 0.631 for NN-skip-10, NN-skip-20, NN-skip-30 and NN-skip-40 respectively) than one from pipeline algorithm. Additionally, the computational requirements for deep neural networks also has been reduced than one for the target pipeline.

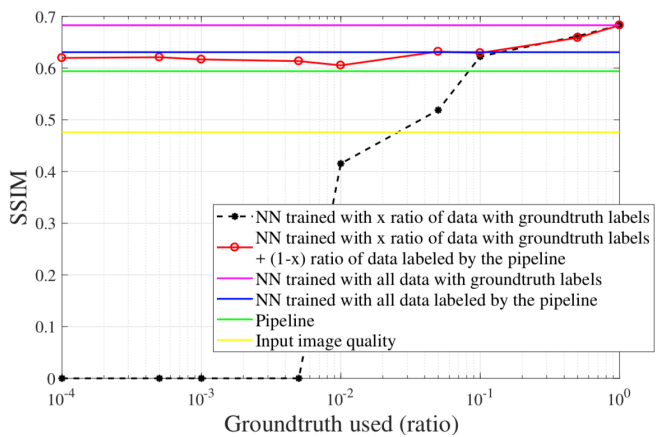

Figure 3: Remodeling a target image denoising pipeline with a deep neural network when partial data with ground truth labels are available: Performances of NN-skip-40 trained only with data ground truth labeled, and then, trained with data with ground truth labels together with data labeled by the target pipeline are compared. We note that we can avoid overfitting due to the data labeled by the target pipeline even when only 40 images (ratio = $10^{-4}$) with ground truth labels are available.

We also simulated a case where partial data with ground truth labels are available and the other data are unlabeled. Figure 3 compares various denoising framework performances when some (x ratio) of data with ground truth labels and the other (1-x ratio) data unlabeled are available. We trained NN-skip-40 only with data with ground truth labels and trained the same network with data with ground truth labels together with data labeled by the target pipeline. We observed that bootstrapping the deep neural network with data from the target pipeline showed better performance than the deep neural networks solely trained with data with ground truth labels. When only 40 images with ground truth labels ($10^{-4}$ ratio) are available, the denoising performance is almost 0 if we train the deep neural network solely with the data with ground truth. If we, however, bootstrap the deep neural network with the data labeled by the target pipeline, the performance is 0.619 which is even better than the existing pipeline (0.594).

### 4.2 Image Classification

For image classification experiments, we have used CIFAR10 [7] data that consists of 60000 32x32 color images in 10 classes, with 6000 images per class. 50000 images are designated for training and the other 10000 images are for testing. To simulate a target pipeline for image classification for CIFAR10 dataset, we made an arbitrary 5 layers of neural network[†]. The deep neural networks used for optimizing the target pipeline are LeNet [8], Net in Net [10], AllNet [17] and DenseNet-40 [4][‡].

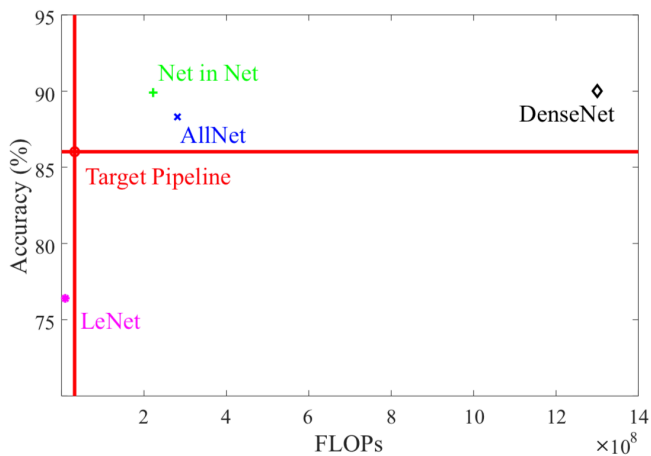

Figure 4: Replacing a target image classification pipeline with deep neural networks: The deep neural networks are trained solely by data labeled by the target pipeline. Some Neural Network present better performance than the target pipeline.

We simulated that there is only an unlabeled data for training the deep neural networks that replace the target pipeline. We first acquire all the parameters for the target pipeline by training the target pipeline using all CIFAR10 training data with ground truth labels. And, we trained the deep neural networks with CIFAR training data labeled by the target pipeline trained. Figure 4 shows pipeline and deep neural networks performances on test data according to computational requirement. The accuracy of the pipeline presents 86% on test data. The performances of Net in Net, AllNet and DenseNet are 89.9%, 88.3% and 90% respectively which are better than the one from target pipeline although Net in Net, AllNet and DenseNet require more computational cost. Thanks to the robustness in the deep neural networks, the deep neural networks trained with noisy labels outperform the target pipeline.

We, here, describe how we simulate a case where a part of data with ground truth labels for training a deep neural network is available for image classification experiments. For this experiment we need two sets of independent data with ground truth labels, one for training the target pipeline to acquire all the parameters, the other one for training deep neural networks. We have selected random 25000 images in training data (refer to data Φ) which is used for training the target pipeline to acquire all the parameters of the target pipeline. The other 25000 images in training data (refer to data Ψ) is used for training deep neural networks.

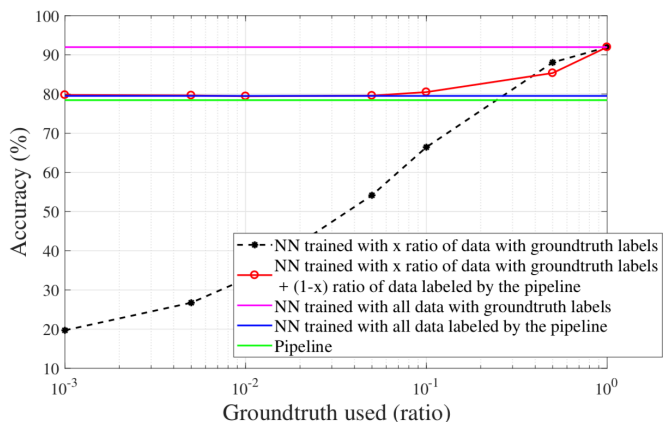

Figure 5: Remodeling a target image classification pipeline with a deep neural network when partial data with ground truth labels are available: DenseNet trained only by data with ground truth labels, and then, trained by data with ground truth labels together with data labeled by the target pipeline are compared. When the deep neural network is bootstrapped by the target pipeline, we can avoid overfitting.

Figure 5 shows various image classification framework performances when some (x ratio) of data Ψ with ground truth labels and the other (1-x ratio) of unlabeled data Ψ are available. Same as to the aforementioned denoising experiment, we trained DenseNet only with x ratio of data Ψ with ground truth labels, and then, trained the same network with x ratio of data Ψ with ground truth labels together with (1-x) ratio of data Ψ labeled by the target pipeline. We observed that the deep neural network bootstrapped with data labeled from the target pipeline showed better performance than the deep neural network solely trained with x ratio of data Ψ with ground truth labels. When very small x ratio, for example $x = 10^{-3}$ (25 images), of data Ψ with ground truth labels are available, if we train the neural network solely with that data, the performance is 19.7% due to overfitting. If we, however, bootstrap the neural network with the (1-x) ratio of data Ψ labeled by the target pipeline, the performance is remained (79.8%) even better than the original pipeline performance (78.4%).

---

[†] Two layers are convolutional neural nets followed by RELU activation function with all 64 output channels. The following three layers are full connection with 384, 192 and 10 number of output neurons respectively.

[‡] DenseNet-40 consists of three dense blocks of 12 layers each. The growth rate is 12.

## 5 Conclusion

In this paper, we have described a workflow that can improve accuracy and reduce computational cost of existing image processing and computer vision pipelines without throughout domain knowledge of the pipelines. Bottleneck parts or all of the target pipeline are identified and then replaced with deep neural networks. In our workflow, we directly address the need for a large corpus of labeled data to train the deep neural networks. Specifically, our workflow uses data labels generated by the target pipeline to bootstrap the training of the deep neural network that is to replace the target pipeline. We experimentally show that (1) we can start with a significantly small amount of ground truth labels and (2) we can avoid overfitting.

Due to the robustness and the ability of deep neural networks to generalize, we observed that the performances of the deep neural networks trained by noisy labels generated by the target pipeline can achieve even better results than the performance of the target pipeline. There is still a number of future efforts to better formalize this workflow. For example, this framework can be applied general signal processing applications, and we propose to benchmark other applications beyond image processing and computer vision functions. We anticipate that there is a spectrum of complex functions that can be easily approximated (e.g. those with well mapped tasks). We have observed that software functions with coding errors (e.g. improperly codified or capabilities that are heuristically disabled) can cause undue noise in the generated dataset. For example, an OpenCV function with a setting to ignore faces that are too small, can generate noisy labels for a deep NN to approximate face detection. As future work, we aim to study noise tolerance for the deep NNs.

Our result addresses machine learning with less manually curated labels and have significance to future design of complex image processing and computer vision systems. In our workflow, we show that we can use a simple target pipeline (e.g. which can be older deep networks such as LeNet or AlexNet of yesteryear) to generate labels and arrive at a larger training dataset. We show that a specified target pipeline can be replaced or approximated with alternative versions in the form of a deep network. Developers can use this workflow to make design tradeoffs (among algorithm performance and computational requirements) that best fit their needs (e.g. lower power consumption and hardware cost for inference).

## ACKNOWLEDGMENTS

This material is based upon work supported by the Office of Naval Research (ONR) under contract N00014-17-C-1011, and NSF #1526399. The opinions, findings and conclusions or recommendations expressed in this material are those of the author and should not necessarily reflect the views of the Office of Naval Research, the Department of Defense or the U.S. Government.